\documentclass[conference]{IEEEtran}
\IEEEoverridecommandlockouts

\usepackage[T1]{fontenc}
\usepackage{courier}

\usepackage{cite}
\usepackage{algorithmic}
\usepackage{graphicx}
\usepackage{textcomp}
\usepackage{xcolor}

\usepackage{microtype}

\usepackage{graphicx}
\usepackage{wrapfig}
\usepackage{booktabs}
\usepackage{multirow}
\usepackage{comment}
\usepackage{enumitem}
\usepackage{tcolorbox}
\tcbuselibrary{breakable}
\usepackage{url}

\usepackage{xr}

\def\BibTeX{{\rm B\kern-.05em{\sc i\kern-.025em b}\kern-.08em
    T\kern-.1667em\lower.7ex\hbox{E}\kern-.125emX}}
\begin{document}

\let\oldtexttt\texttt 
\renewcommand{\texttt}[1]{%
  \oldtexttt{%
    \fontdimen2\font=0.4em\relax 
    #1%
  }%
}

\title{%
  \raisebox{-0.3\height}{\includegraphics[height=2em]{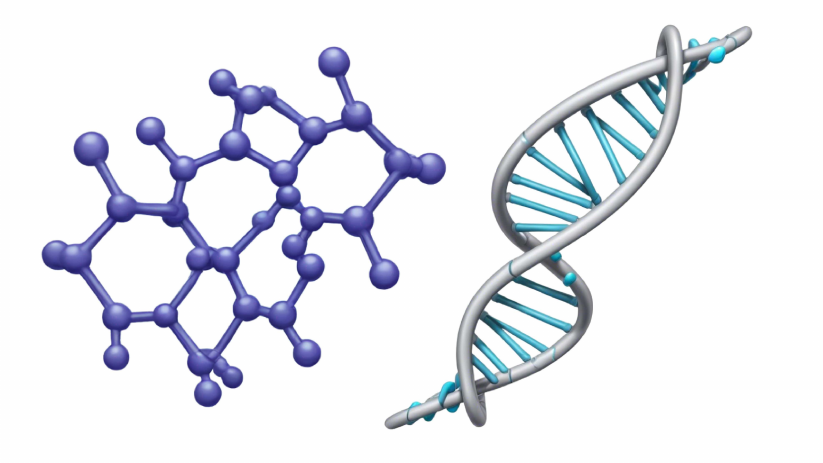}}%
  \hspace{0.1em}%
  {\LARGE\bfseries\texttt{\textcolor{teal}{BioMol-MQA:}}}%
  {\LARGE\bfseries\ A \underline{M}ulti-Modal \underline{Q}uestion \underline{A}nswering Dataset For LLM Reasoning Over \underline{Bio}-\underline{Mol}ecular Interactions}
}

\author{
    \IEEEauthorblockN{Saptarshi Sengupta, Shuhua Yang, Paul Kwong Yu, Fali Wang, Suhang Wang}
    \IEEEauthorblockA{\textit{The Pennsylvania State University}, University Park, PA, USA \\
    \texttt{\{sks6765,sky5341,pky5070,fqw5095,szw494\}}@psu.edu}
}

\maketitle

\begin{abstract}
Retrieval-augmented generation (RAG) has shown success in improving Large Language Model (LLM) performance. However, existing RAG-based LLMs typically retrieve information from a single modality, mainly text. In many real-world settings, such as healthcare, query-relevant information can manifest in various modalities, including knowledge graphs, clinical notes, and complex molecular structures. Thus, being able to retrieve relevant multi-modal domain-specific information and reason over it to generate an accurate response is critical. However, the lack of datasets impedes progress in this direction. To address this gap, we present \texttt{BioMol-MQA}, a new question-answering (QA) dataset on \textit{polypharmacy}, which is composed of two parts: (i) a multi-modal knowledge graph (KG) with text and molecular structure for information retrieval; and (ii) challenging questions that are designed to test LLM capabilities in retrieving and reasoning over the multi-modal KG to answer questions. Our benchmarks indicate that current LLMs struggle to answer these questions and perform well only with the necessary background data, signaling the necessity for strong RAG frameworks. All prompts and evaluation rubrics are provided in our code base.
\textcolor{red}{Disclaimer: This dataset is for LLM research purposes and should not be used to assist with clinical diagnosis.}
\end{abstract}


\section{Introduction}
\label{sec:introduction}

LLMs \cite{zhao2023survey} have shown remarkable performance in various tasks, such as text summarization \cite{zhang2024benchmarking} and question-answering (QA) \cite{singhal2025toward}. Their success has attracted people to adopt them for daily use, e.g., to address medical concerns \cite{mendel2025laypeople}. Despite their strong abilities, they still hallucinate \cite{10.1145/3703155}, face knowledge cutoff \cite{cheng2024dated}, and lack domain knowledge \cite{gu2025effectiveness}. This hinders their adoption in high-stakes settings, such as healthcare, where incorrect responses can have devastating consequences.

\begin{figure}
    \centering
\begin{tcolorbox}[fontupper=\small, fontlower=\small,boxsep=1pt,left=2pt,right=2pt,top=2pt,bottom=2pt]
\textit{DRUG 1: Ketorolac}

\textit{DRUG 1 BACKGROUND: ... is a potent nonsteroidal anti-inflammatory drug (NSAID)
indicated for the \textcolor{magenta}{management of moderate to severe nociceptive pain} ... resulting in the \textcolor{magenta}{attenuation of prostaglandin synthesis} ...}

\textit{DRUG 2: Oxaprozin}

\textit{DRUG 2 BACKGROUND: ... is a nonsteroidal anti-inflammatory drug (NSAID), ... \textcolor{orange}{used to relieve joint pain associated with osteoarthritis and rheumatoid arthritis}. Chemically, it \textcolor{orange}{is a propionic acid derivative}...}

\textit{DRUG-DRUG INTERACTION: Ketorolac-\textcolor{blue}{gastric inflammation}-Oxaprozin}

\tcblower
\textit{Q: Which medication, used for \textcolor{magenta}{moderate to severe nociceptive pain} and known for robust \textcolor{magenta}{inhibition of prostaglandin synthesis}, may contribute to \textcolor{blue}{gastric inflammation when used in combination} with a \textcolor{orange}{propionic acid derivative NSAID indicated for arthritis}?}

\textit{\textbf{\textcolor{cyan!90}{A: Ketorolac}}}
\end{tcolorbox}
\caption{Example from \texttt{BioMol-MQA}. Colors highlight elements from each information source (\textcolor{blue}{Blue} = \textcolor{blue}{Graph}, \textcolor{magenta}{pink}/\textcolor{orange}{orange} = Text) that a model must connect to answer questions. Here, an LLM first needs to identify the correct NSAID (Oxaprozin), determine its outgoing edges in the graph having a gastric inflammation label, and figure out which drug on the other end is used to manage nociceptive pain.}
\label{fig:question_example}
\vspace{-15pt}
\end{figure}

A popular way to address these issues is Retrieval-Augmented Generation (RAG) \cite{fan2024survey}, i.e., retrieving query-relevant context to ground an LLM's response on it for factual accuracy. However, most methods typically retrieve from a single modality \cite{FlashRAG}, e.g., text \cite{trivedi-etal-2023-interleaving}, knowledge graph \cite{edge2024local}, or image \cite{shalev2025imagerag} only. Although initial attempts for multi-modal RAG have been made \cite{abootorabi2025ask}, i.e., retrieval from more than one modality, they mainly focus on general knowledge domains \cite{abootorabi2025ask} that do not require expertise in any particular area. For many real-world problems, e.g., healthcare, information relevant to medical queries can manifest in various modalities such as KGs, text (clinical notes), and complex molecular structures from crystallography. Thus, being able to retrieve the relevant domain information from multiple modalities and reason over diverse knowledge to generate an accurate response is crucial for LLMs. However, work on understanding and developing LLMs with such abilities is limited \cite{alsaad2024multimodal}. One impediment is the lack of datasets supporting this line of research.

To fill this gap, we develop a multi-modal retrieval and reasoning dataset named \texttt{BioMol-MQA} for QA on \textit{polypharmacy}\footnote{\url{https://my.clevelandclinic.org/health/articles/polypharmacy}}, i.e., the concurrent use of multiple medications to address ailments. Polypharmacy is a serious healthcare issue, as unknown drug interactions can exacerbate rather than improve one's health. Given its importance and the extensive domain knowledge required to answer related questions, polypharmacy is the perfect testbed for testing an LLM's multi-modal reasoning abilities. Here, an LLM must utilize background information on drugs, their interaction partners, and molecular-level details to determine the severity of a drug combination. Additionally, it is known that drugs target proteins \footnote{\url{https://www.proteinatlas.org/humanproteome/tissue/druggable}}. Thus, having protein-level data will help it better resolve polypharmacy queries.

\texttt{BioMol-MQA} is composed of two parts: (i) a multi-modal (KG) for information retrieval; and (ii) complex medical questions that require multi-modal RAG and LLM reasoning for QA. The multi-modal KG provides complex and comprehensive domain knowledge from three modalities: a KG of drug-drug and drug-protein interactions, free-text on each entity providing background knowledge, and molecular structure data on drugs, encoded as \texttt{SMILES} strings (\S \ref{sec:SMILES_mol_modality}). Our questions are constructed by integrating information from these three diverse sources. To answer, an LLM must retrieve information from the three modalities to provide an accurate response. To facilitate evaluation and model fine-tuning, each question has a ground truth answer and resources from which the problem was constructed (\S \ref{sec:q_gen}). An example is shown in Figure \ref{fig:question_example}. We formally define the task as,

\textbf{Task Definition} \textit{Given a multi-modal data structure $\mathcal{D(G, T, S)}$ consisting of a knowledge graph $\mathcal{G(N, E)}$ ($\mathcal{N}$ = Nodes (drugs and proteins) and $\mathcal{E}$ = Edges connecting the nodes), a text corpus on the knowledge graph entities $\mathcal{T}$ and molecular structures of drugs as SMILES strings, $\mathcal{S}$, an LLM must utilize at least two modality combinations ($\mathcal{G+T}$ or $\mathcal{G+S}$) to solve a query $\mathcal{Q}$, whose answer is a node in the graph. Mathematically, the task can be described as a function $\mathcal{F: (D, Q) \rightarrow N}$}.

Experimental results on \texttt{BioMol-MQA}  (\S \ref{sec:exp}) show that current LLMs, on their own, are inept at solving these questions. However, when provided with the relevant context, their performance spikes, indicating the necessity to ground reasoning on information via RAG. 
\texttt{BioMol-MQA} also provides data for benchmarking multi-modal retrievers. Additionally, it has a dedicated training, validation, and test split (\S \ref{sec:q_gen}) that can be utilized for fine-tuning models to jointly learn question and modality (graph/text/SMILES) representations useful for downstream tasks as graph link (determining existence of edges between nodes) \cite{li2023evaluating} or molecular property prediction \cite{guo2023can}.

Our \textbf{main contributions} are: (i) We propose \texttt{BioMol-MQA}, a new question answering dataset for retrieval and reasoning over multi-modal contexts; (ii) We propose a synthetic data generation pipeline for augmenting knowledge graphs and creating questions that integrate diverse modalities; (iii) We show the limitations of current retriever models when querying our knowledge source and frontier LLMs in reasoning in complex scenarios as ours.

\section{Related Work}

\textbf{RAG In Medicine} Although most RAG pipelines are tested on general domain data \cite{FlashRAG, abootorabi2025ask}, there are studies exploring RAG in healthcare. \cite{lozano2023clinfo} builds an application for QA over PubMed\footnote{\url{https://pubmed.ncbi.nlm.nih.gov/}} articles. \cite{wang2023retrievalbased} performs molecule synthesis by retrieving similar samples from a labeled database. However, they follow a \textit{unimodal} setup, i.e., retrieve information from a single source. Although \cite{wang2024biobridge} performs QA using a multi-modal biomedical KG, the questions themselves do not integrate each source, essentially collapsing to unimodal retrieval. 

\textbf{RAG Datasets} Current benchmarks for RAG, such as RAGBench \cite{friel2024ragbench} and ChatRAG Bench \cite{liu2024chatqa}, are a combination of existing QA datasets. The original QA datasets are augmented with corpora for RAG testing. That said, we do find two new datasets that directly support RAG, viz., CRAG \cite{yang2024crag} and STaRK \cite{wu2024stark}. CRAG only provides unimodal questions related to the general domain. STaRK combines questions from graph and text modalities to support multi-modal graph retrieval. However, for each KG triple (entity-relation-entity), they consider the background information on one entity and do not \textit{strongly} integrate relationships. This limits question complexity and overall dataset coverage.

\textbf{Molecular QA} Several datasets about QA on molecules are developed, e.g., \texttt{MoleculeQA} \cite{lu-etal-2024-moleculeqa} and \texttt{PubChemQA} \cite{luo2024biomedgpt}. However, questions in both are \textit{template-based}. \texttt{PubChemQA} asks ``\textit{describe this molecule}'' while \texttt{MoleculeQA} has types such as \textit{Which kind of compound does this molecule belong to?}. This heavily limits sample diversity. Further, they do not consider \textit{relationships} between molecules, are restricted to a single modality, and ignore molecule background knowledge.

\texttt{BioMol-MQA} addresses these limitations by effectively combining multiple modalities, utilizing entity relationships, and creating questions that capture all of this. 

\section{Dataset Construction}
\label{sec:dataset_construction}

\begin{figure}
    \centering
    \includegraphics[width=\linewidth]{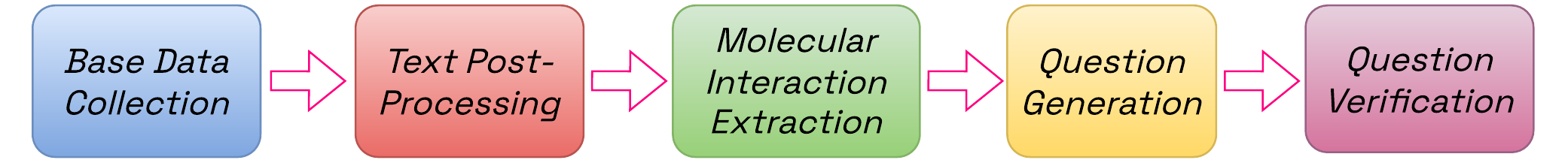}
    \caption{Our dataset development pipeline.}
    \label{fig:complete_pipeline}
    \vspace{-10pt}
\end{figure}
    
The overall pipeline of developing our dataset is summarized in Figure \ref{fig:complete_pipeline}. First, we \textit{assemble our base data} (\S \ref{sec:base_data_acquisition}) where we label our KG and augment its nodes with background text and molecular structures. This transforms the base KG into a multi-modal data structure. Next, we post-process (\S \ref{sec:complexification}) our corpus to enhance question complexity. To address the absence of molecular interactions in the KG, we use an LLM to augment the graph (\S \ref{sec:mol_int_ext}), adding another source for question generation. Next, we create our dataset of LLM-derived questions incorporating each modality (\S \ref{sec:q_gen}). Finally, we assess the quality of our dataset (\S \ref{sec:q_ver}) using a human and an automatic evaluator. 

\subsection{Base Data Acquisition (Stage I)}
\label{sec:base_data_acquisition}

To fully test the ability of existing systems for retrieving and reasoning multi-modal information to answer complex medical queries, our dataset offers three modalities that they must utilize for answering: (i) a \textbf{knowledge-graph} of drugs and proteins, where each node is a drug or protein and each edge is an interaction between two entities; (ii) a \textbf{free-text} associated with each node in the graph, which provides background information such as uses, behavior, etc.; and (iii) the \textbf{molecular structure} of drugs represented by a non-natural language string called SMILES (Simplified Molecular Input Line Entry System) \cite{10.1145/3715318}, which gives insight into the components potentially responsible for participating in various bio-molecular interactions. 

\subsubsection{\textbf{Graph Modality}}
\label{sec:Graph_Modality}

Our base knowledge graph is from \cite{zitnik2018modeling}, consisting of two node types - drugs and proteins, and three edge types, where each edge is an interaction between the corresponding entities. In total, there are 645 unique drugs, 19K proteins, $\sim$4M drug-drug interactions (DDI), $\sim$130K drug-protein interactions (DPI), and $\sim$715K protein-protein interactions (PPI). Only DDIs are labeled ($\sim$1.3K unique) with their polypharmacy side-effects.

\textbf{Resolving Entity Names:} The base KG provides entity/node names in a coded format. Drugs are represented by a \texttt{CID} (Compound Identifier), such as \texttt{CID000002088} (\textit{alendronic acid}). Proteins are described by an NCBI (National Center for Biotechnology Information) Entrez\footnote{\url{https://www.ncbi.nlm.nih.gov/gene}} database entry, such as \texttt{3351} (ID for the gene \texttt{HTR1B}). This ID is mappable to its corresponding protein. These are identifiers used in biomedical research. However, as we are building a natural language QA dataset, we cannot refer to entities by their IDs as they are only meaningful in the context of their respective databases and not scientific discourse. For example, \texttt{3386} (\textit{Fluoxetine}) only makes sense when connected to the PubChem\footnote{\url{https://pubchem.ncbi.nlm.nih.gov/}} database and typically not found in research studies. Moreover, we need to identify the names of those drugs/proteins to search for literature (\S \ref{sec:Background_Info}), as using IDs yields no documents. To address the issues, we obtain the natural language representation of these entities as follows.

\textbf{(A) Resolving Drug Names:} The process of naming a drug is a long and arduous one. After discovery, a potential drug compound is first registered in a medical database, such as PubChem, and then goes through several layers of clinical trials\footnote{\url{https://www.merckmanuals.com/home/drugs/brand-name-and-generic-drugs/overview-of-generic-drugs-and-drug-naming}} to test for efficacy and ultimately marketability, where it gets a generic name such as \textit{Fluoxetine}. Unfortunately, most drugs do not make it through such rigorous testing, which can often take as long as 10 years. Thus, despite being registered, we do not find much information on them. 

\begin{figure}
    \centering
    \includegraphics[scale=0.19]{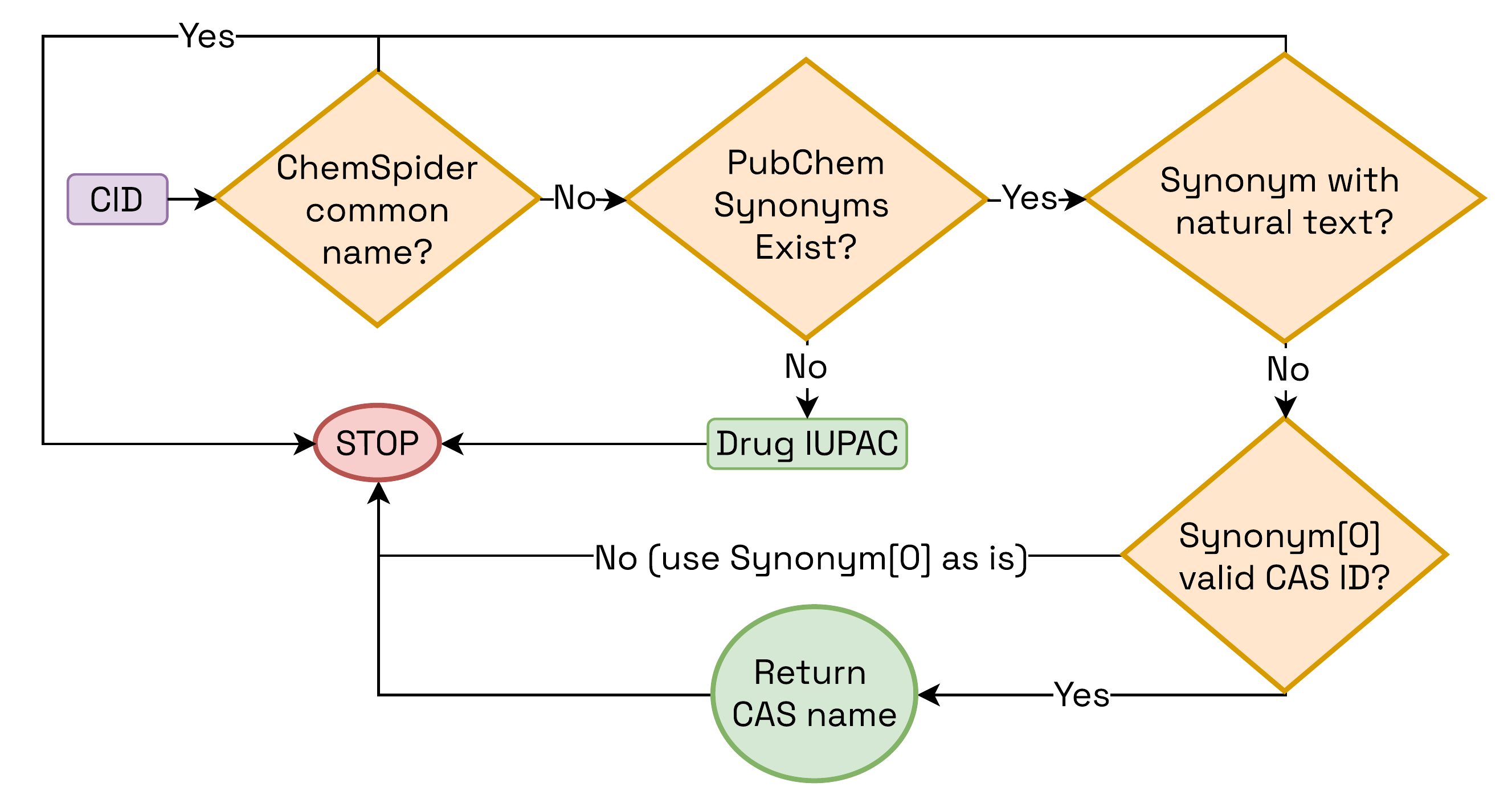}
    \caption{Flowchart for resolving drug names.}
    \label{fig:drug_name_flowchart}
    \vspace{-10pt}
\end{figure}

For our purposes, we require a drug to have been studied and documented for utilization in our dataset. We follow a multi-step process to resolve drug names, as shown in Figure \ref{fig:drug_name_flowchart}, to maximize dataset coverage. A drug is initially represented by a CID or PubChem Compound Identifier, such as \texttt{CID000002088} (\textit{alendronic acid}). First, we query its entry in the PubChem database to obtain its International Chemical Identifier (InChI) key, a unique compound identifier shared across databases. Using this key, we query ChemSpider \footnote{\url{https://www.chemspider.com/}} as its API directly provides a drug's common name. If this fails, we return to PubChem and determine if a set of drug \textit{synonyms} (names used across literature or clinical trials) exists. 

If no synonym exists, we consider the IUPAC (International Union of Pure and Applied Chemistry) name (a long standardized name describing the entire molecular structure) as the drug's name. Otherwise, we try to find any synonym with \textit{natural text}, i.e., no non-alphabet characters (numbers, hyphens, etc.). If there exists no such synonym, then we consider the first synonym in the returned list as the common name, as we observe that, more often than not, the first entry is a reasonably well-structured string. Also, we find that the first synonym might be a CAS\footnote{\url{https://www.cas.org/cas-data/cas-registry}} (Chemical Abstracts Service) number, which gives us a final chance to obtain a common name for the drug.

\textbf{(B) Resolving Protein Names:} Acquiring protein names is more straightforward than acquiring drug names. In the base knowledge-graph, proteins are represented by their source \textit{gene}, molecules that encode information for synthesizing proteins\footnote{\url{https://medlineplus.gov/genetics/understanding/basics/gene/}}. As explained before, a gene is encoded as an NCBI Entrez database entry. Since our goal is to acquire the protein associated with these genes, we query \texttt{STRING}\footnote{\url{https://string-db.org/}} (Search Tool for the Retrieval of Interacting Genes/Proteins), a database tailored for protein interactions. A useful feature of the STRING API is that, given a gene ID, it automatically maps it to the main protein in its database. This helps us resolve each gene to its corresponding protein name. For example, \texttt{HTR1B} (gene) gets mapped to \texttt{5-hydroxytryptamine receptor 1B} (protein).

\begin{wraptable}{r}{3.5cm}
    \caption{Graph Metrics}
    \scriptsize
    \centering
    \begin{tabular}{@{}ll@{}}
    \toprule
    \textbf{Component} & \textbf{Count} \\ \midrule
    Drugs              & 494            \\
    Proteins           & 198            \\
    Drug-Drug Edges    & 18585          \\
    Drug-Protein Edges & 314            \\
    Drugs with Text    & 179            \\
    Proteins with Text & 51             \\
    Drugs with SMILES  & 494            \\ \bottomrule
    \end{tabular}
    \label{tab:graph_stats}
\end{wraptable}

\textbf{Knowledge Graph Sampling:} We do not create questions for every edge, as not all nodes have mineable text data (\S \ref{sec:Background_Info}), which is needed for question generation (\S \ref{sec:q_gen}). Thus, to maintain decent coverage while ignoring unusable nodes, we sample a subgraph as follows (statistics shown in Table \ref{tab:graph_stats})

\textbf{(A) DDI Sampling}: The base KG has $\sim$4M drug-drug interactions. However, as explained, most drugs do not have common or, generic names needed to collect background text for question generation (c.f. \S \ref{sec:q_gen}). As such, we sample a set of DDIs yielding a total of 494 drugs with $\sim$18.5K edges.

\textbf{(B) DPI Sampling/Labeling}: DPIs are unlabeled in the base KG. Without knowing what relationship exists between a drug and a protein, we cannot create questions about them. To address this, we use \texttt{STITCH} (Search Tool For Interactions Of Chemicals) \cite{kuhn2007stitch}, a database integrating information on drugs and their associations with proteins.

We use the subset\footnote{\url{http://stitch.embl.de/download/actions.v5.0/9606.actions.v5.0.tsv.gz}} of STITCH on human DPIs. All DPIs have a \textit{score} (numeric value indicating the strength of the interaction), \textit{mode} (phenotype, etc.), and \textit{action} (activation or inhibition). Simply, \textit{mode} means the type or nature of interaction between the drug and protein, while \textit{action} refers to the outcome of the interaction. It should be noted that \textit{action} is \textit{directional} in nature, i.e., to show which entity acts on which.

For each DPI in our graph, we query the \texttt{STITCH} database to acquire the edge labels. If the DPI does not exist in STITCH, we discard it. If it exists, STITCH returns a (score, mode, action) triple for the DPI. We retain only those edges whose score is greater than 900, i.e.,  indicating an extremely likely or strong interaction. The total number of proteins interacting with each of the 494 drugs is 198, yielding $\sim$15K DPI edges. However, as most of these edges are absent in STITCH, it ultimately leaves a total of 314 DPI edges.

To obtain a unified label, we combine \textit{mode} and \textit{action}, giving us a label set of \textit{binding and inhibition} and \textit{binding and activation}, which indicates that a drug \textit{binds} to a protein and either \textit{inhibits} or \textit{activates} its (proteins) function.

\textbf{(C) PPI Absence}: There are several protein-protein interactions in the base KG. However, like DPIs, they are unlabeled and require us to query the \texttt{STRING} database to get the associated data. When two genes, such as \texttt{114787} and \texttt{2775}, are queried, STRING returns a dictionary of their associated proteins and interactions as \texttt{\{"ascore": 0.083, "tscore": 0.618, "nscore": 0, ...\}}. Each score is a source of evidence for the protein-protein interaction (PPI). For example, \texttt{tscore} (text score) means that a relationship exists between the two proteins as documented by existing studies, etc. Converting these scores to a natural language label raises questions such as which score(s) to choose, how to aggregate them, etc. As there is no straightforward way to acquire a label from these scores, we decided not to include PPIs in our dataset.

Overall, our resulting graph contains $\sim$18.5K DDIs and 314 DPIs, totaling $\sim$18.9 edges, 494 drug and 198 protein nodes.

\subsubsection{\textbf{Text Modality}}
\label{sec:Background_Info}

The second modality is text. To associate each node with a document, we use Wikipedia summaries. We consider sourcing abstracts from PubMed \cite{white2020pubmed}. However, they are usually clinical studies on a \textit{specific} aspect of a drug or protein rather than broader knowledge. Additionally, they are short in length. As such, combining many would lead to an incoherent document for processing. Thus, we use Wikipedia, which provides sufficient and succinct background for an entity. 

We query Wikipedia for each entity in our graph. As explained previously, while many drugs are registered in medical databases, there is no guarantee that we can obtain information beyond basic properties, as most drugs fail clinical trials. As such, we focus on drugs having a generic or commercial name identified in \S \ref{sec:Graph_Modality} to obtain background data easily. This excludes drugs such as \texttt{(+)-Vinblastine} (stereochemistry information), giving a total of 179 drugs and 51 proteins with Wikipedia entries. These entities yield 500 DDIs and 84 DPIs (Table \ref{tab:ques_dist}).

\subsubsection{\textbf{SMILES}}
\label{sec:SMILES_mol_modality}

Our final modality is the molecular structures of drugs. Drugs are ultimately compounds with distinct behaviors and structures. As such, there is valuable information to be gleaned by studying their molecular composition, such as reaction sites and functional groups, which motivates us to include drug structure as our third modality. SMILES is a popular notation used to represent chemical compounds in electronic databases. It provides a set of rules to define molecular structure in a \textit{non-natural} string. For example, \texttt{C=C} indicates \texttt{Ethene} (\texttt{CH\textsubscript{2}CH\textsubscript{2}}) and \texttt{=} refers to a double bond between the carbon atoms. Using these rules, drugs and compounds are standardized. Proteins do not have a SMILES representation as they are not ``small molecules'' like drugs \cite{lin2019bigsmiles}. Thus, we can only use the molecular structure of drugs in our dataset. We obtain SMILES for each drug in the KG by querying PubChem, a database of chemical compounds. 

\subsection{Text Post-Processing (Stage II)}
\label{sec:complexification}

\begin{wraptable}{r}{4cm}
    \scriptsize
    \caption{Post-processing (PP) outcomes. Numbers represent the average over the documents.}
    \vskip -1.2em
    \centering
    \begin{tabular}{@{}lcc@{}}
    \toprule
    & \textbf{Before PP} & \textbf{After PP} \\
    \midrule
    Tokens      & 300 & 233 \\
    Readability & 15  & 28  \\
    \bottomrule
    \end{tabular}
    \label{tab:complexification}
\end{wraptable}

Most Wikipedia summaries are condensed and have the relevant biomedical data we need. However, a summary may also include irrelevant information such as market statistics and historical data. Additionally, many summaries are verbose and can inundate an LLM's context window \cite{liu2024lost}. To address these issues, we utilize GPT's paraphrasing abilities \cite{hassanipour2024ability}. First, we identify documents having more than 200 tokens (empirically). This excludes all protein texts as they have an average of 114 tokens, while for drugs, it is 212. Next, we prompt GPT-4o \cite{hurst2024gpt} to rewrite the source text, ignoring historical data and using domain-specific jargon suitable for an expert audience. There are two reasons for doing this: (i) to enhance question complexity based on the rephrased text; and (ii) to reduce their chance of \textit{cheating} by using direct references from Wikipedia, as most LLMs have extensive training on it \cite{johnson-etal-2024-wikimedia}.

\begin{figure}
    \centering
    \includegraphics[width=\columnwidth]{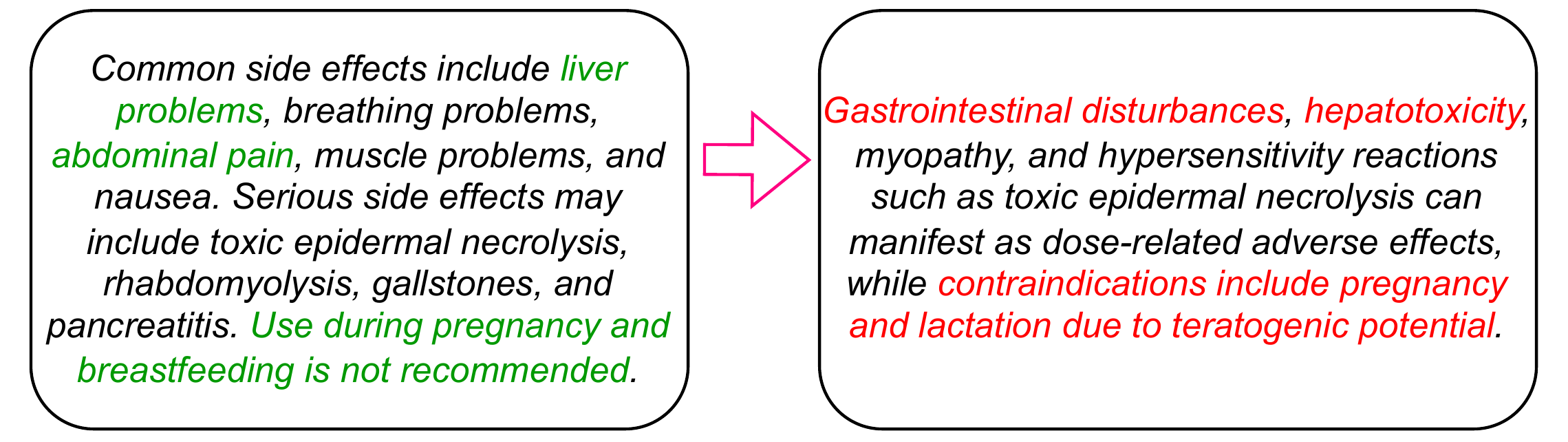}
    \caption{Simple phrases as \textit{liver problems} are replaced by technical jargon as \textit{hepatotoxicity} by post-processing.}
    \label{fig:complexification}
\end{figure}

We assess the outcome of this process in two ways: (i) \textit{average tokens} (conciseness) and (ii) \textit{readability scores} (text complexity). For readability, we use \texttt{Gunning-Fog} Index \cite{gunningfog}, which indicates the level of expertise required to comprehend a given text. Higher values reflect content suitable for audiences with formal training in a discipline. Results from the process are in Table \ref{tab:complexification}. As seen, the enhanced texts are more condensed, with readability almost doubled. This achieves our goal of having a semantically dense corpus addressing the above issues. Fig. \ref{fig:complexification} shows the effect on \textit{Fenofibrate}'s (drug) text.

\subsection{Molecular Interactions (Stage III)}
\label{sec:mol_int_ext}

The relationships described in the base KG are at the \textit{physio/biological level}, i.e., physical manifestations~of taking two drugs (e.g., \textit{nausea}, etc.) or influence of a drug on protein activity (\textit{binding}, etc.). However, there are interactions at the \textit{molecular} level that describe chemical associations between the atoms of molecules. Drugs are small molecules that also exhibit such interactions. Unfortunately, these are not provided by the base KG. Ignoring them would eliminate insights obtained from a crucial aspect of DDI.

To remedy the absence of molecular interactions, we leverage GPT-4o's knowledge of chemistry. As shown by prior studies \cite{hatakeyama2023prompt}, LLMs have strong capabilities in reasoning through chemistry tasks, including molecular property prediction from SMILES analysis \cite{guo2023can}. This lends credence to its ability to extract molecular interactions. Given their SMILES, GPT-4o is asked to describe potential associations between the atoms of two drugs. An LLM's interpretable nature provides insight into its reasoning process for such a complex task. 

\begin{figure}
    \includegraphics[width=\linewidth]{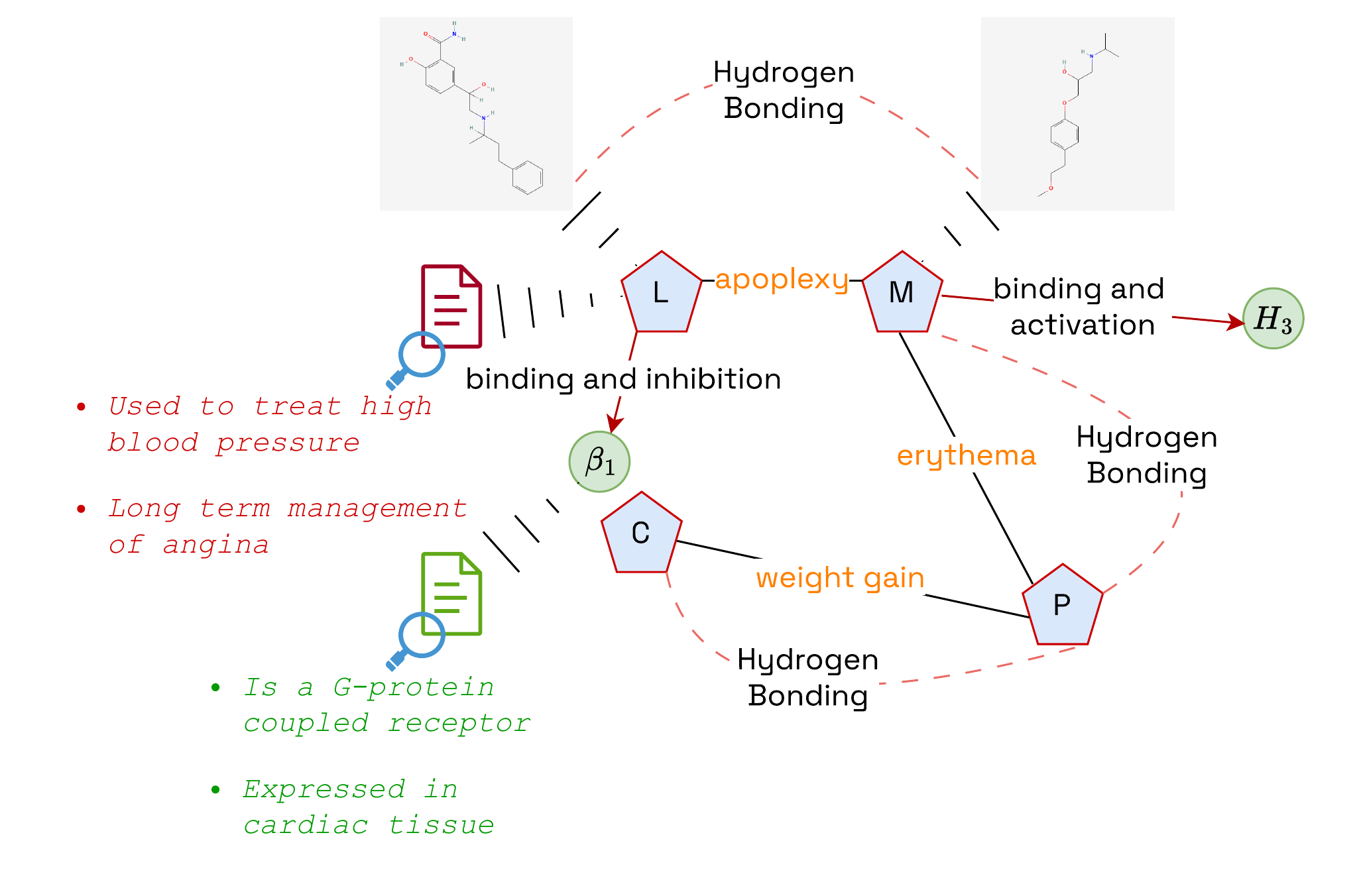}
    \vskip -1em
    \caption{The final knowledge-graph depicting each modality, edge direction, and augmented edges (molecular interactions). Entities in \textcolor{blue}{blue} are drugs, and those in \textcolor{green}{green} are proteins. Dashed edges indicate molecular interactions, and solid edges are DDI/DPI, depending on the type of entities. L = \textit{Labetalol}, M = \textit{Metoprolol}, $\beta_1$ = \textit{Beta-1 adrenergic receptor}, $H_3$ = \textit{Histamine H3 receptor}, P = \textit{Pirbuterol}, C = \textit{Carvedilol}}
    \label{fig:final_graph}
\end{figure}

We have GPT output four fields, (i) \textit{interaction name}: a label for the potential interaction (ii) \textit{mechanism}: why it occurs, (iii) \textit{evidence}: cues from the SMILES to lend credence to its reasoning and (iv) \textit{severity}: how strong the interaction is (low/medium/high), if it can be identified. Of the 500 drug-drug pairs that we use for question-generation (c.f. \ref{sec:Background_Info}), GPT can extract molecular relationships for 499 pairs. Investigating the generations reveals five types of relationships: \texttt{Hydrogen Bonding}, \texttt{$\pi$-$\pi$ Stacking}, \texttt{Steric Clashes}, \texttt{Ionic}, and \texttt{Electrostatic Interaction}. The introduction of these interactions transforms the base graph into a \textit{multi-graph}, i.e., a graph with multiple edges between two nodes, adding a further layer of nuance. Fig. \ref{fig:final_graph} depicts a portion of the final graph.

\subsection{Question Generation (Stage IV)}
\label{sec:q_gen}

LLMs have demonstrated great performance in rewriting or generating desired questions that follow a given prompt and context \cite{abdullin-etal-2023-synthetic, wu-etal-2024-synthetic, hemati-beigy-2024-consistency, liu2024synthetic}. Leveraging these capabilities, we \textit{prompt} GPT-4.1 \cite{openaiIntroducingGPT41} for our question generation. For a knowledge graph triple (entity-relation-entity), GPT is either given background text on each entity or their SMILES (for drugs). It is then asked to create a question by integrating the relational information with the background data, such that the answer is one of the respective KG nodes. The idea is that, \textbf{to answer the questions, a model must utilize multiple information sources (graph/text/molecular structure) to provide an accurate response}. Each question has a single gold-truth answer (one of the nodes associated with the respective KG triple) and context needed to answer it.

Our questions are not multiple-choice, i.e., we do not provide models with options to choose from. This is a design choice to reflect real-world information retrieval (IR) settings \cite{abbasiantaeb2021text, biancofiore2024interactive} wherein users query databases without associated choices. Questions based on DDIs are of two types according to the relationship (physio/biological and molecular), while there is only one type of question for a DPI. To craft questions, GPT is always provided with the knowledge-graph interaction triples between the respective entities. Additionally, it receives background text on each entity (for Bio-based DDIs and DPIs) or SMILES strings (for molecular DDIs). The answer for a DDI is always a drug, while for a DPI, it is the protein, as we wanted to capture the directional nature of the interaction.

The above questions are \textit{single-hop}, i.e., utilize properties of two connected entities. To add a further layer of complexity, we also create \textit{multi-hop} reasoning questions, i.e., those involving two entities connected by a path longer than 1. For this, we randomly sample 100 two- and three-hop DDI and DPI edges. The overall distribution of questions is in Table \ref{tab:ques_dist}. We divide the questions in a stratified 80-10-10 ratio into training/validation/test, resulting in 1346, 169, and 168 samples in the respective splits. For each question, we have the ground-truth answer and context (KG triples/entity background text/SMILES) used to construct the question.

\begin{table}[t]
\centering \small
\caption{Question distribution in the dataset.}
\label{tab:ques_dist}

\setlength{\tabcolsep}{4pt}
\renewcommand{\arraystretch}{0.95}
\begin{tabular}{@{}llcccc@{}}
\toprule
\textbf{Entity Pair} & \textbf{Interaction Type} & \textbf{1-hop} & \textbf{2-hop} & \textbf{3-hop} & \textbf{Total} \\
\midrule
Drug - Protein       & DPI                        & 84  & 100 & 100 & 284 \\
\multirow{2}{*}{Drug - Drug} 
                     & Physiological (Bio) & 500 & 100 & 100 & 700 \\
                     & Molecular     & 499 & 100 & 100 & 699 \\
\midrule
\end{tabular}
\vspace{-15pt}
\end{table}

We formulate two key requirements for crafting questions: (i) \textbf{Entity names must not appear verbatim in the question} - To push a model to understand the context, connect it to the knowledge base, and then infer the entity name as opposed to directly learning the entities involved. This aligns with our goal of probing the multi-modal information processing abilities of LLMs; and (ii) \textbf{Questions should always test for the relationship between entities rather than isolated facts on each} - As our objective is to incorporate relational (graph) data and frame questions around it, instead of developing straightforward factoid questions.

For criterion one, we check for the presence of the entities used to create the respective question. Overall, we do not find any major leakage, thus affirming our first criterion. The only exception is that 15\% of DPI questions mention the protein name. We excuse the LLM for doing this as proteins, overall, do not have much background data for it to utilize. Except that, none of the questions mention drug names or leak their SMILES strings, showing that the model followed our instructions well.

For criterion two, we notice that most questions have identifiable phrases to indicate a relationship between the entities. For example \textit{combined with}, \textit{associated with}, etc. We use the proportion of questions having these phrases as a rough proxy for our relationship criteria. This yields a $\sim$73\% hit rate, indicating that the majority of questions follow our instructions. As it is difficult to automatically evaluate the other questions due to variations in phrasing, holistic examination indicates that all questions have a relational component from the KG.

\begin{wraptable}{r}{5cm}
\caption{Question Statistics.}
\label{tab:q_stats}
\scriptsize
\centering
\begin{tabular}{@{}ll@{}}
    \toprule
    \textbf{Metric}           & \textbf{Value} \\ \midrule
    Question Length (Tokens)    & 66.6           \\
    Type-to-Token ratio (TTR)   & 0.84           \\
    Shannon Entropy             & 5.67           \\
    Dependency Parse Tree Depth & 10.86          \\ \bottomrule
    \end{tabular}
\end{wraptable}

\textbf{Quantitative metrics} include, Question Length, Type-to-Token Ratio (number of unique words/total words), Entropy (how much information is conveyed by the text), and dependency parse tree depth (a structure to describe grammatical roles and relationships between words). Scores in Table \ref{tab:q_stats} reflect the linguistic and semantic complexities of our dataset. 

\textbf{Qualitative analysis} by our human evaluator (\S \ref{sec:q_ver}) reveals that our \textit{1-hop questions are harder} to answer than the 2/3-hop ones. This counterintuitive observation is explained by the fact that the latter category of questions has higher information density, or more content to learn from. Overall, their assessment rated the questions as fair and suitable for advanced students preparing for the USMLE (exam to get medical license).


\subsection{Question Verification (Stage V)}
\label{sec:q_ver}

We assess question quality by (i) \textbf{Automatic Verification}. Using LLMs to assess data quality (\texttt{LLM-as-a-judge}) \cite{gu2024survey, li2024generation, li2024llms} has emerged as a promising proxy for human evaluation. By describing a set of grading criteria (rubric), frontier LLMs \cite{achiam2023gpt, hurst2024gpt} can provide assessments aligning with real annotators \cite{gu2024survey, li2024generation, li2024llms}. However, using the same LLM for generation and evaluation can introduce biases \cite{NEURIPS2024_7f1f0218}. Thus, we use \texttt{Claude-3.7 Sonnet} \cite{anthropicClaudeSystem} for evaluation; (ii) \textbf{Human Evaluation}. Despite the proficiency of \texttt{LLM-as-a-judge}, they are prone to issues such as preferring verbose outputs and prompt sensitivity \cite{koo-etal-2024-benchmarking}. Thus, having a human-in-the-loop + judge-LLM is effective for gauging data quality. One of our co-authors, a PhD student in Bioinformatics, provides the domain knowledge required to evaluate the questions.

\begin{wraptable}{r}{5cm}
\caption{Human vs. Automatic Question Eval.}
\label{tab:eval}
\scriptsize
\centering
\begin{tabular}{@{}lcc@{}}
\toprule
\textbf{Metric} & \textbf{Human} & \textbf{Automatic (LLM)} \\
\midrule
Clarity & 1.22 & 1.75 \\
Coverage & 1.79 & 1.97 \\
Assumptions & 1.79 & 1.98 \\
Inference & 1.79 & 1.91 \\
\bottomrule
\end{tabular}
\end{wraptable}

We randomly sample 100 QA pairs for the LLM and human annotator to provide feedback. Both are given a rubric of 4 custom metrics: \textit{Clarity} (technical prowess needed to understand the question); \textit{Coverage} (proportion of modalities utilized); \textit{Assumptions} (including information beyond what was supplied (hallucination)); \textit{Inference} (can the answer be derived by studying the provided data \cite{jullien-etal-2024-semeval}). Each metric is scored as 0, 1, or 2, depending on the guidelines. Averaged scores for each metric for human and LLM evaluation are in Table \ref{tab:eval}. As we see, there exists an overall agreement between our human evaluator and the LLM. The only measure where the two differ is in \textit{Clarity}, with the LLM recognizing that the questions on average are harder (higher scores = greater complexity). This makes sense as \texttt{Claude} is a generalist model \cite{guan2024towards} while our human annotator is a well-read medical student, conversant with the language and content of the questions. 

\section{Experiments}
\label{sec:exp}


To validate the quality of \texttt{BioMol-MQA}, we perform two categories of experiments (i) \textit{LLM-reasoning} and (ii) \textit{retrieval}-based generation. We aim to show that (i) LLMs \textit{can} answer \textit{some} of our questions, but struggle with others due to a lack of domain knowledge (\S \ref{sec:llm_reasoning}) and, (ii) RAG helps, but it is still far-away from the upper bound performance (\S \ref{sec:retrieval_tests})

\subsection{LLM Reasoning Capability} 
\label{sec:llm_reasoning}

\begin{table*}[t]
\centering
\caption{Reasoning benchmark scores (without RAG). \textbf{Bold} represents the best performing model in each category for the corresponding test. Each tuple is (lexical EM, lexical F1, BERTScore F1)}
\label{tab:reasoning_bechmark}
\resizebox{\textwidth}{!}{%
\begin{tabular}{@{}c|cc|ccccc@{}}
\toprule
\textbf{Model}    & \multicolumn{2}{c|}{\textbf{Closed Source LLMs}} & \multicolumn{5}{c}{\textbf{Open Source LLMs}} \\ \midrule
\textbf{Approach} & o4-mini & Claude 3.7 Sonnet & DeepSeek R1 & Llama 3.3 & Qwen 3 & TxGemma & Mistral \\
Zero-Shot &
  \textbf{(0.32, 0.42, 0.79)} &
  (0.30, 0.36, 0.78) &
  \textbf{(0.33, 0.40, 0.87)} &
  (0.27, 0.34, 0.85) &
  (0.22, 0.25, 0.51) &
  (0.07, 0.16, 0.82) &
  (0.02, 0.14, 0.81) \\
Upper Bound &
  (0.88, 0.88, 0.94) &
  \textbf{(0.89, 0.89, 0.94)} &
  \textbf{(0.90, 0.90, 0.98)} &
  (0.71, 0.71, 0.93) &
  (0.80, 0.80, 0.88) &
  (0.76, 0.76, 0.94) &
  (0.74, 0.76, 0.95) \\ \bottomrule
\end{tabular}%
}
\end{table*}

Here, we investigate LLM reasoning under (i) zero-shot, i.e., without \texttt{RAG} (lower bound) and (ii) with the gold data (KG-triples, background text/SMILES) (upper bound). 

\textbf{LLMs Used} We benchmark seven models on the test split of our dataset, which include: (i) Two closed-source LLMs, \cite{anthropicClaudeSystem}  and \texttt{o4-mini}\footnote{\url{https://developers.openai.com/api/docs/models/o4-mini}}; and (ii) five open-source LLMs, \texttt{TxGemma} \cite{wang2025txgemma} (trained to understand information on various areas of drug development), \texttt{DeepSeek-R1} \cite{guo2025deepseek}, \texttt{Llama 3.3}\footnote{\url{https://www.llama.com/docs/model-cards-and-prompt-formats/llama3_3/}}, \texttt{Qwen3} \cite{yang2025qwen3}, and \texttt{Mistral} \cite{jiang2023mistral7b}. Each model is first asked to provide their thought process to arrive at the answer and then the answer itself, a strategy found to improve LLM performance \cite{wang-etal-2024-large-language-models-fair, yugeswardeenoo-etal-2024-question}. We use \textit{Together AI}\footnote{\url{https://www.together.ai/}} for the best variants of open-source LLMs (App. \ref{sec:LLM_settings}).

\textbf{Metrics} LLM answers are evaluated using (i) Lexical EM (exact match): 0 (miss)/1 (hit) measure of string equality, (ii) Lexical F1: Balanced proportion of token overlap between the predicted and reference strings and (iii) \texttt{BERTScore} F1 \cite{bert-score}: An embedding-based similarity measure between two strings. Using lexical and embedding-based measures allows a holistic evaluation of LLM performance and dataset quality.

\textbf{Analysis} Table \ref{tab:reasoning_bechmark} shows the results. We observe: \textbf{(i)} The average zero-shot EM/F1/BERTScore (across all models) is 0.22, 0.28, and 0.77, respectively. This clearly shows that existing LLMs are not yet equipped to handle these types of questions by themselves as they lack the necessary domain knowledge; \textbf{(ii)} The average EM/F1/BERTScore (across all models) for the upper-bound test is 0.81, 0.81, and 0.94, respectively. This indicates two things, \textbf{(1)} \textit{the questions in our dataset, while challenging, are answerable and fair as supported by these scores}, which demonstrate the quality of our data; \textbf{(2)} the importance of grounding answers through RAG, i.e., \textit{LLMs are capable of handling these questions, provided they are given the necessary background information}, which further show the necessity of our multimodal KG. For example, models such as Claude and DeepSeek almost triple their EM while Mistral's F1 improves fivefold. 

\subsection{Retriever Performance / Multimodal RAG}
\label{sec:retrieval_tests}

Here, we examine how well current retrievers can query our knowledge sources. We test three types of retrievers, text-only, graph-only, and a combination of both as our multi-modal baseline, which simply combines the best text/graph retriever.

\textbf{Retrievers Used} For text-only retrieval, we test \textit{sparse}: \texttt{BM25} \cite{schutze2008introduction} and the following \textit{dense} models, 

\begin{itemize}[leftmargin=*]
    \item \texttt{MedCPT} \cite{jin2023medcpt}: Tailored for medical text by extensively tuning a \texttt{PubMedBERT} \cite{gu2021domain} checkpoint for query-article retrieval.
    \item \texttt{MolLM} \cite{mollm}: Fine-tuned BERT \cite{devlin-etal-2019-bert} on text, \texttt{SMILES}, and molecular graphs, providing insights for \texttt{SMILES} retrieval.
    \item \texttt{DPR} \cite{karpukhin-etal-2020-dense}: BERT-based retriever trained for QA tasks.
    \item \texttt{text-embedding-3-large}\footnote{\url{https://developers.openai.com/api/docs/models/text-embedding-3-large}} is our only proprietary model. As such, we do not have training details and can only use its reported results as evidence for its utility. 
\end{itemize}

For our graph-only retriever, we implement a simple \texttt{Neo4J}\footnote{\url{https://neo4j.com/}} baseline. It works as follows,
\begin{enumerate}[leftmargin=*]
    \item It first builds an index, i.e., a graph database using the provided set of triples (entity-relation-entity).
    \item Next, it creates embeddings (using \texttt{all-MiniLM-L6-v2} \cite{huggingfaceSentencetransformersallMiniLML6v2Hugging}) for each relation in the database to locate relationships semantically similar to the one described by the question.
\end{enumerate}

For a given query,
\begin{enumerate}[leftmargin=*]
    \item It first extracts entities in the question via Named Entity Recognition (NER) using Flair\footnote{\url{https://flairnlp.github.io/docs/tutorial-embeddings/flair-embeddings}}) models.
    \item Next, it looks for relations closest to the query using the same sentence encoder used to embed the relations.
    \item It returns the top-k triples with these entities/relationships.
\end{enumerate}

We tried different settings (using a simpler regex-NER, different embedding models such as PubMedBERT \cite{gu2021domain}, etc.) and landed on this version, which provides decent results. Although a simple baseline, we find it to be reasonably effective. That said, given the graph's complexity, we only use the set of triples for question creation (\S \ref{sec:q_gen}) for retrieval.

\textbf{Metrics} Retrieval accuracy is measured using three metrics, (i) Hits@k - (0/1) measure to see if \textit{any} (soft hits) or \textit{all} (hard hits) relevant items are among the top-k retrieved results; (ii) Recall@5 - The proportion of correct items among the top-5 retrieved results (iii) MRR (Mean Reciprocal Rank) - Inverse index of the first relevant result.

\begin{table}[t]
\centering
\caption{Retriever performances. The first 5 rows are text-only retrievers. SMILES information is encoded as text and falls under text-only retrieval. *Neo4j was run on the subset of triples used for question generation due to poor performance on the entire graph. \textbf{Hits are provided as hard/soft}. Hybrid = {[}BM25 + Neo4j{]}}
\label{tab:retriever_performance}
\begin{tabular}{@{}cccccc@{}}
\toprule
\textbf{Retriever}     & \textbf{Hit@5}    & \textbf{Hit@10}    & \textbf{Hit@15}   & \textbf{Recall@5} & \textbf{MRR}  \\ \midrule
BM25                   & \textbf{0.2/0.58} & \textbf{0.28/0.58} & \textbf{0.3/0.58} & \textbf{0.4}      & \textbf{0.52} \\
DPR                                        & 0/0.14    & 0.01/0.23 & 0.06/0.27 & 0.06 & 0.07 \\
MedCPT                                     & 0.10/0.49  & 0.15/0.56 & 0.21/0.57 & 0.27 & 0.45 \\
MolLM                                      & 0.01/0.38 & 0.03/0.47 & 0.04/0.51 & 0.17 & 0.24 \\
\texttt{TE3L} & 0.15/0.57         & 0.24/0.58          & 0.28/0.58         & 0.35              & 0.49          \\ \midrule
Neo4j*                                     & 0.15/0.19 & 0.19/0.26 & 0.2/0.28  & 0.05 & 0.13 \\ \midrule
Hybrid & 0.06/\textbf{0.60}      & 0.13/\textbf{0.60} & 0.14/\textbf{0.61} & 0.32      & \textbf{0.52}     \\ \bottomrule
\end{tabular}%
\end{table}

\textbf{Analysis} Results from benchmarking retrievers are in Table \ref{tab:retriever_performance}. We observe: \textbf{(i)} For text-only retrievers, a simple BM25 baseline outperforms \textit{each} dense embedding model. This is supported by studies \cite{sciavolino-etal-2021-simple} which show how lexical models such as BM25 can generalize across domains, occasionally outperforming dense embeddings. \textbf{(ii)} For our graph-only retriever, we notice a similar trend, where a basic database retriever (Neo4j) outperforms trained dense models (\S App. \ref{sec:GNN}). As we explain in \S App. \ref{sec:GNN}, training GNNs on our graph \textit{directly} is difficult due to its small size (KGs typically have millions of nodes and edges \cite{ji2021survey}) and complexity. Thus, using simpler baselines (Neo4j) is preferable in settings such as ours; and \textbf{(iii)} Our hybrid retriever (BM25+Neo4j) shows a balanced performance between the text-only and graph-only modes. Although overall hard hit decreases, soft hit is quite strong, outperforming each modality individually. This indicates the benefit of using even a simple multi-modal retrieval baseline.

\begin{figure}
    \centering
    \includegraphics[width=\columnwidth]{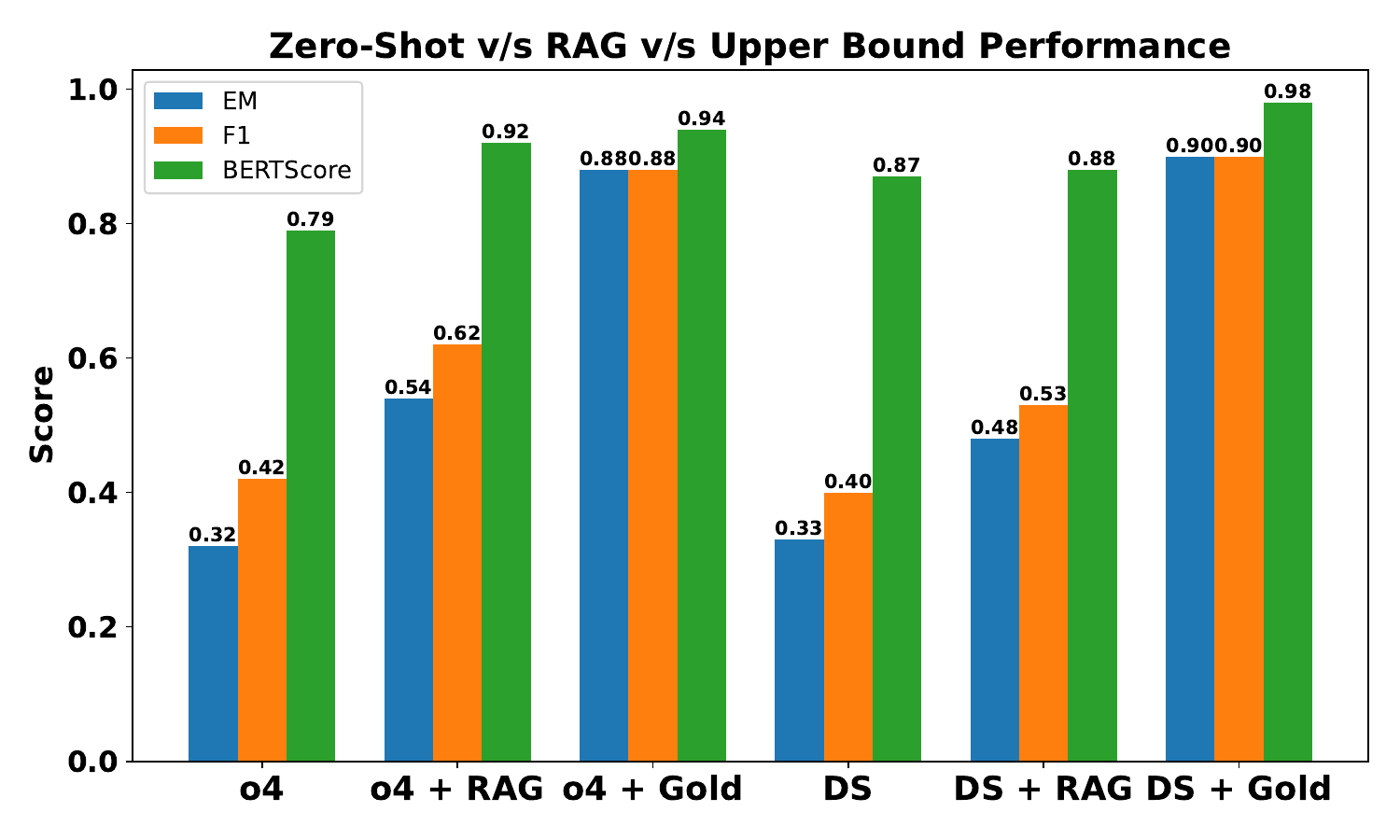}
    \caption{Impact of RAG on o4-mini and DeepSeek R1 (DS). For comparison, their zero-shot and upper bound (with the gold data) performances are provided.} 
    \label{fig:rag}
\end{figure}

\textbf{Multimodal RAG Performance} To test how well our models perform using RAG, we use our hybrid retriever (BM25/Text + Neo4j/Graph) with the two best zero-shot models, i.e., \texttt{o4-mini} and \texttt{DeepSeek-R1}. This experiment not only establishes the importance of grounding LLM responses on appropriate background knowledge but also the limitations of existing retrievers in accessing complex data sources such as ours. Results from this test are provided in Fig. \ref{fig:rag}. The scores highlight three things: (i) the importance of using even a basic retriever to provide domain knowledge to LLMs, (ii) poor performance of current retrievers in processing and providing multi-modal data to LLMs, and (iii) the gap remaining between RAG-based reasoning and an LLM's true potential.

\section{Conclusion and Future Work}

We present \texttt{BioMol-MQA}, a multi-modal reasoning and retrieval benchmark for polypharmacy queries. Through our experiments, we show the limitations of existing retrievers and models in dealing with multi-modal contexts such as ours. With our dataset, the goal is to provide a testbed for evaluating models for handling complex, real-world medical queries. We will continue to grow the dataset, such as including \textit{unanswerable} questions, exploring modalities such as protein structures (long-chain amino acid sequence) for testing the length limits of LLMs, and multi-agent workflows to enhance question diversity.

\section*{Acknowledgment}
This work is supported by, or in part by, National Science Foundation award \#2555559.

\appendices

\section{LLM Settings}
\label{sec:LLM_settings}

We run inference on our test set with 7 LLMs (o4-mini, Claude 3.7 Sonnet, DeepSeek R1, Llama 3.3 (Llama-3.1-8B-Instruct-Turbo), Qwen 3 (Qwen/Qwen3-235B-A22B-fp8-tput), TxGemma (TxGemma-9B-predict), Mistral (Mistral-7B-Instruct-v0.2). Each model is run with default settings as provided by the respective APIs. TxGemma (API unavailable) / dense retrievers use a single NVIDIA A6000 48GB GPU.

\section{Graph Retrievers}
\label{sec:GNN}

Training graph neural network (GNN) retrievers is not as straightforward as text-based retrievers due to the ingrained differences in structure (edges, features, etc.) across graphs \cite{hou2024graphalign}. That said, we attempt to train three GNN-retrievers, viz., Convolutional Network (GCN), SAGE, and Attention Network. We utilize the entire graph (c.f. Table \ref{tab:graph_stats}) along with the questions in the training split of our dataset and evaluate on the validation split. The idea here is to train a GNN to retrieve triples (entity-relation-entity) by jointly optimizing triple and query embeddings. Unfortunately, due to the size of our training data and graph, we are unable to learn useful representations as GNNs are usually trained on much larger samples \cite{snapnets}. Table \ref{tab:gnn_full} shows the results. 

\begin{table}[ht]
\centering
\caption{Neo4j + GNN retrievers trained on the entire graph. Scores are based on the test set.}
\label{tab:gnn_full}
\small
\begin{tabular}{cccc}
\hline
\textbf{Model}            & \textbf{Hits@15} & \textbf{Recall@5} & \textbf{MRR} \\ \hline
Neo4j                     & 0.07/0.09        & 0.01              & 0.03         \\
Graph Convolution Network & 0.03/0.06        & 0.01              & 0.01         \\
Graph SAGE Network        & 0.01/0.06        & 0                 & 0.01         \\
Graph Attention Network   & 0.04/0.09        & 0.01              & 0.02         \\ \hline
\end{tabular}
\end{table}

Due to the size of the graph, we treat it as a homogeneous structure (single entity/relationship type), similar to the base graph learning method from \cite{zitnik2018modeling}. Treating our knowledge graph as a heterogeneous graph (multiple node/edge types) and using RGCN (Relational-GCN) layers did not work. This makes sense as RGCN networks typically work well with larger data and a higher number of edges per relationship \cite{thanapalasingam2022relational}. We try to \textit{hot start} our optimization by pretraining a GNN for a \textit{link prediction} (detecting if an edge exists between nodes). However, this also yielded similar results as above.

After hyperparameter exploration, we settled on each GNN being a two-layer network, with ReLU activation. They were trained to minimize triplet loss, i.e., given a triple of (query, positive triple, negative triple), a model learns to minimize the distance between the query and positive triple (from our ground truth set) while maximizing the distance between the query and negative triple. We train our models for 10 epochs with an AdamW optimizer and a learning rate of $1e-4$. We use \texttt{all-MiniLM-L6-v2} \cite{huggingfaceSentencetransformersallMiniLML6v2Hugging} as our query encoder.

\section*{Resource Availability}
Our code and dataset are available at \texttt{\url{https://github.com/saptarshi059/biomolqa}} and \texttt{\url{https://huggingface.co/datasets/BioMolMQA/BioMolMQA}} respectively.

\bibliographystyle{IEEEtran}
\bibliography{references}

\end{document}